\documentclass{ica}
\usepackage[round,authoryear,comma]{natbib}
\usepackage{mathptmx}
\usepackage[T1]{fontenc}
\usepackage{setspace}
\usepackage{geometry} 
\usepackage{epstopdf}
\usepackage[labelsep=period]{caption} 
\usepackage{subcaption}
\usepackage{graphicx}
\usepackage{amsmath}
\usepackage{cleveref}
\usepackage{amssymb}
\usepackage{booktabs}
\usepackage{multirow}
\usepackage[table]{xcolor}
\DeclareMathOperator*{\argmin}{arg\,min}
\DeclareMathOperator*{\argmax}{arg\,max}
\usepackage{fancyhdr}

\newcommand{\eg}{\textit{e.g.,}}

\fancypagestyle{first}{%
    \fancyhead[R]{}
    
}

\begin{document}
\title{Semantic Segmentation for Sequential Historical Maps by Learning from Only One Map}

\author{
 Yunshuang Yuan\,\textsuperscript{a}\thanks{Corresponding author} , Frank Thiemann\,\textsuperscript{a}, Monika Sester\,\textsuperscript{a}}

\address{
      \textsuperscript{a }Leibniz University Hannover, [firstname].[lastname]@ikg.uni-hannover.de\\
}
\sloppy

\abstract{
Historical maps are valuable resources that capture detailed geographical information from the past. However, these maps are typically available in printed formats, which are not conducive to modern computer-based analyses. Digitizing these maps into a machine-readable format enables efficient computational analysis. In this paper, we propose an automated approach to digitization using deep-learning-based semantic segmentation, which assigns a semantic label to each pixel in scanned historical maps. A key challenge in this process is the lack of ground-truth annotations required for training deep neural networks, as manual labeling is time-consuming and labor-intensive. To address this issue, we introduce a weakly-supervised age-tracing strategy for model fine-tuning. This approach exploits the similarity in appearance and land-use patterns between historical maps from neighboring time periods to guide the training process. Specifically, model predictions for one map are utilized as pseudo-labels for training on maps from adjacent time periods.
Experiments conducted on our newly curated \textit{Hameln} dataset demonstrate that the proposed age-tracing strategy significantly enhances segmentation performance compared to baseline models. In the best-case scenario, the mean Intersection over Union (mIoU) achieved 77.3\%, reflecting an improvement of approximately 20\% over baseline methods. Additionally, the fine-tuned model achieved an average overall accuracy of 97\%, highlighting the effectiveness of our approach for digitizing historical maps.
}

\keywords{historical maps, semantic segmentation, weakly supervision}

\maketitle

\thispagestyle{first}


\section{Introduction}\label{intro}

\begin{figure*}[t!]
    \centering
    \includegraphics[width=.8\linewidth]{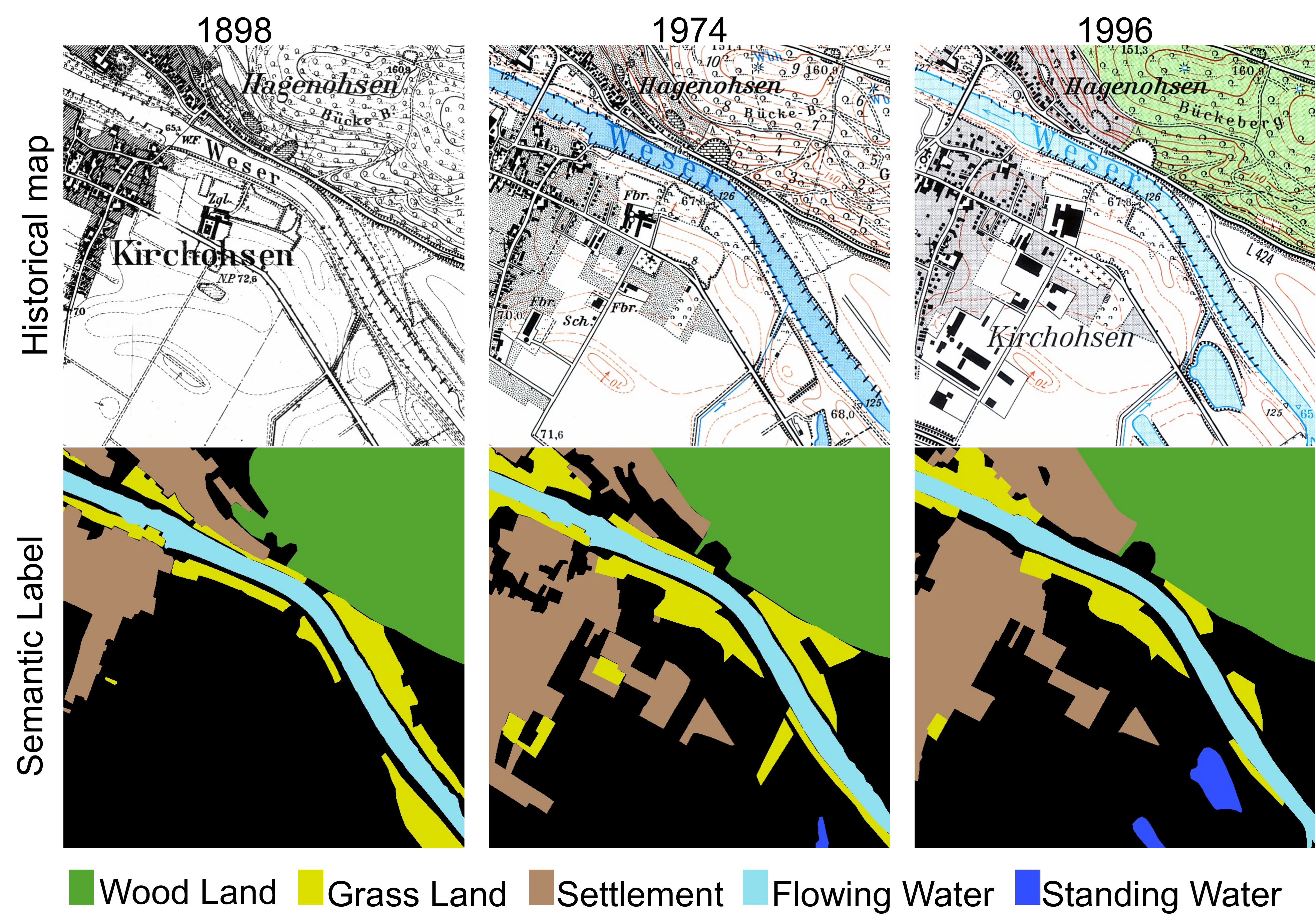}
    \caption{An example of historical maps and the corresponding labels of Hameln, Germany.}
    \label{fig:hameln}
\end{figure*}
Historical maps store the geographical knowledge about the past. They provide valuable insights into changes of urban development and land use over time. Despite their significance, these maps are originally in analog or scanned digital forms, limiting their accessibility and usability for contemporary research and applications. Extracting and interpreting the information contained in historical maps is crucial for advancing spatio-temporal analyses of land use development. 

Conventional methods for this task often rely on geographic information systems (GIS) to interactively and manually digitize geographical features or regions into vectorized polylines or polygons~\citep{salt_marsg,Levin_Kark_Galilee_2010,san-antonio-gomez_urban_2014,picuno_investigating_2019}. These geometric representations are then converted back into raster or pixel formats to facilitate quantitative analyses, such as assessing habitat changes~\citep{salt_marsg}, urban development~\citep{san-antonio-gomez_urban_2014} or hydrogeomorphological changes~\citep{tonolla_seven_2021}.
However, manual labeling is a tedious and time-consuming process. To enhance efficiency, previous studies~\citep{leyk_segmentation_2010,Uhl_Leyk_2021,uhl_automated_2020} have explored methods to extract target pixels (\eg~settlement) by analyzing pixel-level features in historical maps. These features are typically derived from color statistics and spatial patterns in the pixel's neighborhood. Such features, often referred to as hand-crafted, are designed by human experts using statistical and mathematical algorithms. While effective for specific types of historical maps, these algorithms are inherently limited in adaptability. Once tailored to a particular map style, they are difficult to generalize to maps with different designs, restricting their scalability for large-scale applications.

Semantic image segmentation~\citep{Csurka2023SemanticIS,Yuan_gevbev2023} using deep learning models offers a promising approach to automate this process at scale. This technique classifies each pixel in a map image into predefined categories, such as settlements, rivers, or forests. Previous studies have demonstrated the effectiveness of semantic segmentation models for extracting specific features such as building footprints~\citep{heitzler_cartographic_2020}, road networks~\citep{ekim_automatic_2021}, and hydrological features~\citep{wu_closer_2022,wu_leveraging_2022}.
To the best of our knowledge, no prior work has investigated multi-class semantic segmentation to extract pixels representing multiple semantic categories with distinct patterns in historical maps. In this study, we aim to train a semantic segmentation model capable of identifying five classes—\textit{woodland, grassland, settlement, flowing water} and \textit{standing water}—from historical maps of Hameln, Germany (\cref{fig:hameln}). The maps span a temporal range from 1897 to 2017.

Unlike modern maps, which are vectorized into geometric shapes, historical maps—such as those shown in \cref{fig:hameln}—are characterized by intricate cartographic styles and inconsistent labeling. Additionally, they often exhibit artifacts caused by aging, including stains, fading, physical damage, or distortions introduced during digital scanning. These factors pose significant challenges for training a single model capable of generalizing across heterogeneous maps spanning long time periods.
To address the impracticality of manually labeling maps for all time periods, we propose training a generalized model by leveraging a single labeled map from a specific year, referred to as \textit{anchor year}. Specifically, we assume that maps from similar time periods exhibit a high degree of consistency in cartographic style and land-use representation. Based on this assumption, we use a model $\mathcal{M}_a$, trained on labeled data from the anchor year $a$, to generate pseudo-labels for maps within a small temporal range ($a \pm \sigma$, where $\sigma < 10$). These pseudo-labels are subsequently used to fine-tune the model. By iteratively applying this process to data from earlier and later years, the model incrementally learns from maps across the entire temporal range. We term this fine-tuning strategy, which exploits temporal consistency for progressive learning, as \textit{age-tracing}.

In summary, the contributions of this paper are:
\begin{itemize}
    \item we proposed \textit{age-tracing} strategy which leverages the temporal consistency of historical maps for fine-tuning the semantic segmentation network. 
    \item we curated a new dataset \textit{Hameln} for weakly-supervised multi-class semantic segmentation of historical maps.
    \item The experiment results demonstrate that the proposed \textit{age-tracing} strategy significantly enhances the performance of semantic segmentation models using labels from only a single anchor year.
\end{itemize}

\begin{figure*}[t]
    \centering
    \includegraphics[width=0.9\linewidth]{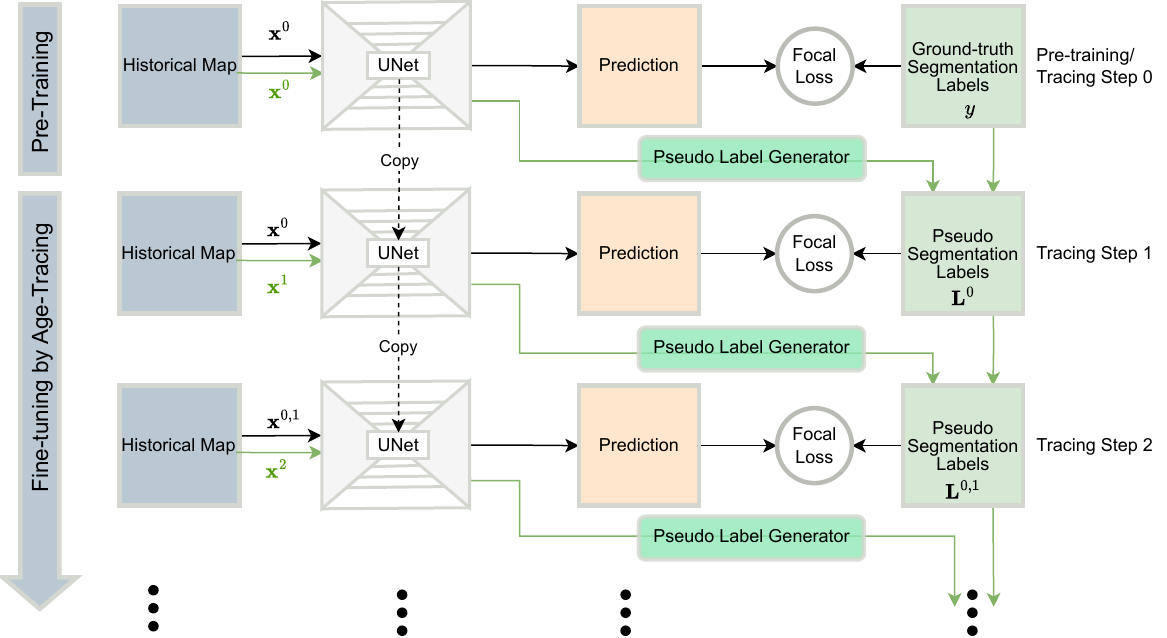}
    \caption{Pipeline for training the \textit{UNet} by age-tracing.}
    \label{fig:pipeline}
\end{figure*}

\section{Related Work}
\subsection{GIS-based Historical map analysis}
Geographic Information System (GIS) offers advanced, feature-rich tools with interactive user interfaces that facilitate the processing and analysis of maps. These tools enable the conversion of raster data into vector-based maps and vice versa, supporting diverse geo-spatial analysis needs.
For instance, \citet{salt_marsg} manually identified and extracted the salt marsh areas from the digitized historical maps photographs for analyzing the salt marsh losses in New England; \citet{Levin_Kark_Galilee_2010} utilized GIS to extract settlement information for analyzing the sedentarization process in southern Palestine; \citet{san-antonio-gomez_urban_2014} employed GIS tools to geo-reference historical and contemporary maps, enabling the manual observation and analysis of urban development in Real Sitio de Aranjuez (Spain); furthermore, \citet{picuno_investigating_2019} investigated the temporal evolution of the rural landscape with GIS, a process that involved manually digitizing land-use classes into polygons and then converting them back into raster maps for further analysis.
These GIS-based analyses, while effective, are highly time-intensive due to the extensive manual labeling requirement, making them impractical for large-scale land-use studies.

\subsection{Unsupervised methods by hand-crafted features}

Hand-crafted features, which capture the local appearance of a pixel, have been widely used in image analysis, particularly in scenarios where specific, non-learnable algorithms rely on color statistics and spatial patterns in the pixel’s neighborhood. These methods are typically unsupervised, requiring no ground truth labels for training, and are designed to automatically identify pixels that conform to predefined feature patterns.

For example, \citet{leyk_segmentation_2010} proposed a method for identifying and segmenting homogeneous regions based on color similarity. Similarly, \citet{Uhl_Leyk_2021} observed that urban areas in historical maps often exhibit highly uniform colors. By analyzing pixel colors in the color space, they applied the k-means clustering algorithm to extract urban regions from such maps.
In contrast, \citet{uhl_automated_2020} trained a convolutional neural network (CNN) to extract building and settlement areas. To reduce the manual effort involved in labeling training data for the CNN, they employed texture descriptors such as Local Binary Patterns (LBP)~\citep{lbp} to generate hand-crafted features for classifying the small patches ($48 \times 48$ pixels) of the historical maps into predefined classes, \eg~building and settlement areas. 
While this approach reduced manual intervention, it limited the CNN’s ability to leverage its full potential. Specifically, the small patch-based input design constrained the model from capturing more complex features across larger receptive fields, which are critical for effectively analyzing historical maps.

Despite their advantages, methods based on hand-crafted features have significant limitations in handling variations in map styles, colors, and patterns. They are particularly sensitive to color noise and inconsistencies introduced by diverse cartographic conventions. For instance, in \cref{fig:hameln}, the river in the map of 1898 is nearly indistinguishable from the background, as both are rendered in white. Furthermore, woodland representation varies substantially between 1898 (scattered symbols) and 2017 (symbols with green color). Such differences in visual styles underscore the challenges of applying hand-crafted methods to historical maps with heterogeneous designs.

\subsection{Semantic segmentation}

Semantic segmentation involves assigning semantic classes to individual pixels in an image, enabling a detailed understanding of its content and context~\citep{Csurka2023SemanticIS}. Early approaches primarily relied on algorithms utilizing hand-crafted features; however, the advent of deep learning has increasingly supplanted these methods.

A pioneering work in deep learning-based semantic segmentation is Fully Convolutional Networks (FCN)~\citep{Shelhamer2014FullyCN}, which employs stacked convolutional layers to transform RGB images into semantic label maps. Building on this foundation, \citet{unet} introduced a U-shaped architecture, known as \textit{UNet}, which incorporates shortcut connections between down-sampling and up-sampling branches. This design aggregates features at multiple resolutions, effectively combining global and local information to enhance the model’s reasoning ability for accurate segmentation. \textit{UNet} has been widely adopted for processing historical maps.
For example, \citet{heitzler_cartographic_2020} employed an ensemble of \textit{UNet} models to extract building footprints from Siegfried maps in Switzerland. Similarly, \citet{ekim_automatic_2021} used \textit{UNet} for road extraction, while \citet{wu_closer_2022,wu_leveraging_2022} applied it to extract hydrological features. However, these methods rely on fully supervised training, necessitating time-consuming and labor-intensive manual labeling processes.

To address this challenge and leverage partial labels for training a single model applicable to maps from diverse time periods, \citet{wu_domain_2023} proposed a domain adaptation~\citep{da} technique for segmenting historical maps. This method introduced a co-occurrence detection module into the \textit{UNet} framework, which generates co-occurrence masks that allow learning only over the masked areas.
In contrast, our work leverages the similarities between maps from adjacent years with available labels, using these relationships to train models with reduced reliance on extensive manual labeling.


\section{Semantic segmentation framework}

\subsection{Task formalization}
A sequence of historical maps is defined as $\mathcal{H}=\{(m_i, a_i)|i\in \{1, \dots,N\}\}$, where $m_i$ is the available historical maps of year $a_i$.
The maps are sorted according to the order $a_{i-1}<a_{i}<a_{i+1}$ and the gap year is $\delta = a_i-a_{i-1}>0$. 
Given the ground truth land use labels $y$ in year $a_y$, we aim to train a general model $\mathbf{M}$ that segments all maps in $\mathcal{H}$ and assigns these segments with one of the semantic labels: \textit{Woodland (WL), Grassland (GL), Settlement (SM), Flowing Water (FW), Standing Water (SW), Unknown (UK)}.

\subsection{Training by age-tracing}

As illustrated in \cref{fig:pipeline}, we employ a \textit{UNet} to process historical maps and generate corresponding predictions, namely the semantic labels for each pixel in the maps. The training process comprises two main stages: pre-training and fine-tuning via age-tracing. In the pre-training stage, the \textit{UNet} is trained using the available ground-truth labels $y$ and historical maps from the anchor age $a_i$, which are the closest to the age of the labeled data. Subsequently, the pre-trained model undergoes fine-tuning through the age-tracing process. During each age-tracing step, additional data from the nearest past and future ages are incorporated into the training set. Pseudo labels for these newly included maps are generated using the model trained in the previous tracing step, allowing the network to progressively learn from a broader temporal range.

\textbf{UNet}: The architecture consists of four down-sampling and four up-sampling blocks. Each down-sampling block comprises two convolutional layers followed by a max-pooling layer, which progressively reduces spatial dimensions while capturing high-level features. Conversely, each up-sampling block includes three convolutional layers, with the first being an up-sampling transposed convolution, which restores spatial dimensions while refining feature maps for accurate pixel-wise segmentation.

\textbf{Pre-training}:  The \textit{UNet} is initially trained using the ground truth labels $y$ of year $a_y$ and the historical maps $X^0$ from the closest year $a_i$ relative to $a_y$, where $i=\argmin_i |a_y-a_i|$. If $a_i\neq a_y$, the pre-trained model $\mathbf{M}^0$ is used to generate the pseudo labels $l_i$ for $a_i$. Otherwise, we set $l_i=y$.  

\textbf{Fine-tuning}: As shown in \cref{fig:pipeline}, the tracing step 1 starts at $a_i$ with the input data $X^0=\{x_i\}$ and the pseudo labels $L^0=\{l_i\}$, where $x_i=(m_i, a_i)$. The model trained in this step is used to generate pseudo labels $L^1=\{l_{i-1}, l_{i+1}\}$ for the additional training data $X^1=\{x_{i-1}, x_{i+1}\}$ in the next tracing step, where $x_*=(m_{*}, a_{*})$. In the tracing step 2, the input maps $X^{0,1}=\{x_{i-1}, x_i, x_{i+1}\}$ and the corresponding labels $L^{0,1}=\{l_{i-1}, l_i, l_{i+1}\}$ are used for fine-tuning. The tracing process steps further with
\begin{align}
    X^{0,\dots,n}&=\{x_{i-n}, x_{i-(n-1)}, \dots, x_{i+(n-1)}, x_{i+n}\}\\
    L^{0,\dots,n}&=\{l_{i-n}, l_{i-(n-1)}, \dots, l_{i+(n-1)}, l_{i+n}\}
\end{align}
until it reaches both ends of the sequential ages.

\subsection{Pseudo-label generation}
Given the classification confidences of one pixel $\textbf{s}=\{s_c|c\in \{0, 1, 2, 3, 4, 5\}\}$, where $c$ is the class six labels including the unknown class. We use the entropy to quantify the uncertainty
\begin{equation}
    u=-\sum_c s_c\log s_c
\end{equation}
for this pixel. We then select an uncertainty threshold $\epsilon$ for generating pseudo labels with
\begin{equation}
    l = \begin{cases}
        \argmax_c (\textbf{s})  &\text{  if } \max(\mathbf{s}) < \epsilon\\
        -1  &\text{  otherwise}
    \end{cases}
\end{equation}
where $-1$ indicates that the corresponding sample pixel is not counted for calculating the loss.

\begin{table*}[ht]
\centering
\begin{minipage}{0.5\textwidth}
\caption{1898}
\centering
\resizebox{\textwidth}{!}{%
    \begin{tabular}{l|c|c|c|c|c|c|c}
    \toprule
    \multirow{2}{*}{Model} & \multicolumn{5}{c|}{IoU} & \multirow{2}{*}{mIoU} & \multirow{2}{*}{OA}\\\cline{2-6}
     & WL & GL & SM & FW & SW &  \\\hline
    $Pre_{bi}$ & 96.6 & 76.8 & 55.3 & \textbf{56.8} & - & 71.4 & 95.9\\
    $All_{bi}$ & 96.6 & 75.1 & 56.7 & 54.1 & - & 70.6 & 95.8\\
    $Trace_{bi}$ & \textbf{96.9} & \textbf{83.2} & \textbf{58.5} & 49.7 & - & \textbf{72.1} & \textbf{96.7}\\
    \hline
    $Pre_{mono}$ & 0.0 & 0.0 & 2.0 & 0.0 & - & 0.5 & 23.4 \\
    $All_{mono}$ & 95.3 & 19.7 & 43.6 & 52.2 & - & 52.7 & 90.6 \\
    $Trace_{mono}$ & \textbf{96.2} & \textbf{42.0} & 35.1 & 46.9 & - & \textbf{55.0} & \textbf{93.2} \\\bottomrule
    \end{tabular}
\label{tab:1898}
}%
\end{minipage}%
\hfill
\begin{minipage}{0.5\textwidth}
\caption{1974}
\centering
\resizebox{\textwidth}{!}{ 
    \begin{tabular}{l|c|c|c|c|c|c|c}
    \toprule
    \multirow{2}{*}{Model} & \multicolumn{5}{c|}{IoU} & \multirow{2}{*}{mIoU} & \multirow{2}{*}{OA}\\\cline{2-6}
     & WL & GL & SM & FW & SW &  \\\hline
    $Pre_{bi}$ & 97.2 & 80.6 & 85.4 & 66.2 & 28.7 & 71.6 & 96.3 \\
    $All_{bi}$ & 97.0 & 79.9 & 86.9 & 64.3 & 28.5 & 71.3 & 96.3 \\
    $Trace_{bi}$ & \textbf{97.4} & \textbf{85.2} & \textbf{90.8} & \textbf{67.9} & \textbf{54.5} & \textbf{79.2} & \textbf{97.3} \\
    \hline
    $Pre_{mono}$ & 0.0 & 0.0 & 11.6 & 50.5 & 11.5 & 14.7 & 49.4 \\
    $All_{mono}$ & 94.2 & 33.5 & 56.9 & 65.7 & 48.3 & 59.7 & 88.6 \\
    $Trace_{mono}$ & \textbf{95.6} & \textbf{54.4} & 50.6 & 61.1 & 27.9 & 57.9 & \textbf{90.0} \\\bottomrule
    \end{tabular}
\label{tab:1974}
}
\end{minipage}

\begin{minipage}{0.5\textwidth}
\caption{1982}
\centering
\resizebox{\textwidth}{!}{%
    \begin{tabular}{l|c|c|c|c|c|c|c}
    \toprule
    \multirow{2}{*}{Model} & \multicolumn{5}{c|}{IoU} & \multirow{2}{*}{mIoU} & \multirow{2}{*}{OA}\\\cline{2-6}
     & WL & GL & SM & FW & SW &  \\\hline
    $Pre_{bi}$ & \textbf{97.7} & 77.2 & 86.3 & 62.6 & 15.3 & 67.8 & 96.4 \\
    $All_{bi}$ & \textbf{97.7} & 75.8 & 87.3 & 60.3 & 15.1 & 67.2 & 96.4 \\
    $Trace_{bi}$ & 97.6 & \textbf{86.4} & \textbf{88.2} & \textbf{64.2} & \textbf{56.8} & \textbf{78.6} & \textbf{97.4} \\
    \hline
    $Pre_{mono}$ & 97.0 & 0.0 & 20.7 & 48.9 & 9.1 & 35.1 & 73.8 \\
    $All_{mono}$ & 96.1 & 33.6 & 56.6 & 62.0 & 60.3 & 61.7 & 90.4 \\
    $Trace_{mono}$ & \textbf{97.6} & \textbf{47.8} & \textbf{56.9} & 60.7 & 33.9 & 59.4 & \textbf{91.3} \\\bottomrule
    \end{tabular}
\label{tab:1982}
}%
\end{minipage}%
\hfill
\begin{minipage}{0.5\textwidth}
\caption{1996}
\centering
\resizebox{\textwidth}{!}{ 
    \begin{tabular}{l|c|c|c|c|c|c|c}
    \toprule
    \multirow{2}{*}{Model} & \multicolumn{5}{c|}{IoU} & \multirow{2}{*}{mIoU} & \multirow{2}{*}{OA}\\\cline{2-6}
     & WL & GL & SM & FW & SW &  \\\hline
    $Pre_{bi}$ & \textbf{97.5} & 70.8 & 87.2 & 57.1 & 8.5 & 64.2 & 95.8 \\
    $All_{bi}$ & \textbf{97.5} & 69.9 & 73.7 & 52.0 & 7.5 & 60.1 & 94.7 \\
    $Trace_{bi}$ & 97.3 & \textbf{81.3} & \textbf{87.7} & \textbf{58.7} & \textbf{61.4} & \textbf{77.3} & \textbf{97.0} \\
    \hline
    $Pre_{mono}$ & 26.6 & 0.0 & 21.5 & 39.1 & 29.8 & 23.4 & 56.3\\
    $All_{mono}$ & 96.3 & 41.0 & 63.1 & 60.6 & 68.7 & 65.9 & 91.8 \\
    $Trace_{mono}$ & \textbf{97.8} & \textbf{50.6} & \textbf{64.1} & 58.8 & 42.1 & 62.7 & \textbf{92.5} \\\bottomrule
    \end{tabular}
\label{tab:1996}
}
\end{minipage}
\end{table*}

\begin{table}[t]
    \centering
    \caption{Available Ground-truth labels.}
    \label{tab:gt_label}
    \begin{tabular}{c|l}\hline
        3821 &  1974, 2023\\\hline
        3822 &  1975, 2023\\\hline
        3921 &  1973, 2023\\\hline
        3922 & 1898, 1974, 1982, 1996 \\\hline
    \end{tabular}%
\end{table}

\section{Experiment}
\subsection{Dataset}
To evaluate the age-tracing training strategy, we introduce the dataset \textit{Hameln}. 
The original map sheets of \textit{Hameln} were produced by the Lower Saxony mapping agency (LGLN). These maps were scanned, color-corrected, and manually georeferenced using four corner points and a control point at the center of each map sheet. A projective transformation was applied to rectify the map borders. The rectified maps were uniformly projected into the UTM coordinate system (EPSG 25832) with a pixel resolution of one meter. The scanned map sheets consist of four patches, labeled 3821, 3822, 3921, and 3922, at a scale of 1:25,000, covering the period from 1897 to 2017. 
For the training process, we use the patches 3821, 3822, and 3921, while patch 3922 is reserved for evaluation. The ground-truth labels are manually annotated (except for the 2023 digital topographic map), as listed in \cref{tab:gt_label}.

\subsection{Experiment configuration}
According to the availability of labels, we design two settings for comparative experiments. In the first setting, we use the labels from the year 1973 to 1975 for training. Note that each training patch in this setting only contains label $y$ from one year $a_y$ ($1973\leq a_y\leq 1975$). In each tracing step, we trace one year further in the past and one in the future as an additional data year for fine-tuning. Because of this bi-directional tracing feature, we note this experimental setting as $Trace_{bi}$. In the second setting, the labels from 2023 are used for pre-training. Since the historical maps are only available from the year 1897 to 2017, the age-tracing process can only trace monotonically to the past, hence the notation $Trace_{mono}$. 

To validate that the age-tracing strategy is beneficial in improving the segmentation performance, we train two additional models for each setting of $Trace_{bi}$ and $Trace_{mono}$. The first model is the pre-training model which is only trained on the map from $a_i$ and ground-truth from $a_y$. This model is notated as $Pre_{bi}$ and $Pre_{mono}$ respectively for the two settings. The second model is trained based on all maps across all ages with regard to the label from $a_y$ as the unified ground-truth for these maps. This model is notated as $All_{bi}$ and $All_{mono}$, respectively. 
\subsection{Training details}
The models $Pre_{bi}$, $Pre_{mono}$  and $All_{bi}$, $All_{mono}$ are trained for 20 epochs with the initial learning rate of $lr=1e-4$. The learning rate is reduced two times at epochs 10 and 15, each with a factor of $0.1$.
$Trace_{bi}$ and $Trace_{mono}$ are fine-tuned for 5 epochs in each tracing step with the initial learning rate of $lr=1e-5$, which is reduced at epoch 3 by factor $0.1$. All models are optimized with the \textit{Adam} optimizer. The weight decay is set to $0.01$. During training, maps are cropped into images of size $384\times 384$ pixels with the overlapping margin of $128$ pixels. Each image is augmented with random flipping along $x$ and $y$ axes and random rotation. During the testing, the maps are cropped into a size of $1024\time 1024$ pixels.

\subsection{Evaluation metrics}
We use two metrics to evaluate the performance of the trained model: the Intersection over Union (IoU) and the Overall Accuracy (OA). Given the prediction $P=\{p_i| i\in X\}$ and the ground-truth labels $Y=\{y_i | i\in X\}$, where $X$ is the set of pixels of the input historical map, the IoU for a class $c$ is calculated with
\begin{equation}
    IoU_c = \frac{|\{i|p_i=c\} \cap \{i|y_i=c\}|}{|\{i|p_i=c\} \cup \{i|y_i=c\}|}
\end{equation}
To have an overall evaluation for all classes, a mean IoU (mIoU) is calculated by averaging the $IoU_c$ of all classes. In addition, the OA is calculated with
\begin{equation}
    OA = \frac{|\{i|p_i=y_i\}|}{|X|}
\end{equation}

\begin{figure*}[t]
    \centering
    \includegraphics[width=\linewidth]{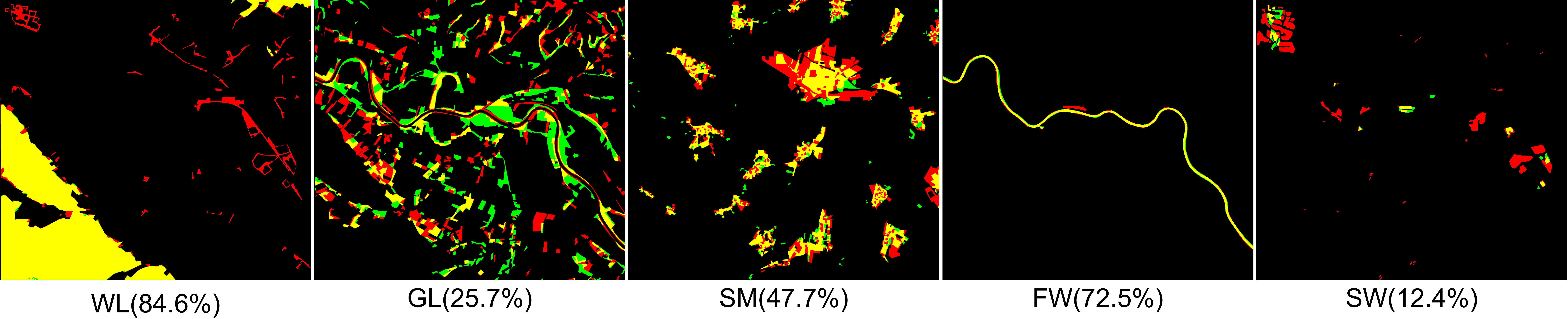}
    \caption{Comparison of labels from 1974 (green) and 2023 (red). The overlapping areas (yellow) are the consistent labels. The percentage values in the brackets are the IoUs between the labels from the two ages.}
    \label{fig:incos}
\end{figure*}
\begin{figure*}
    \centering
    \includegraphics[width=\linewidth]{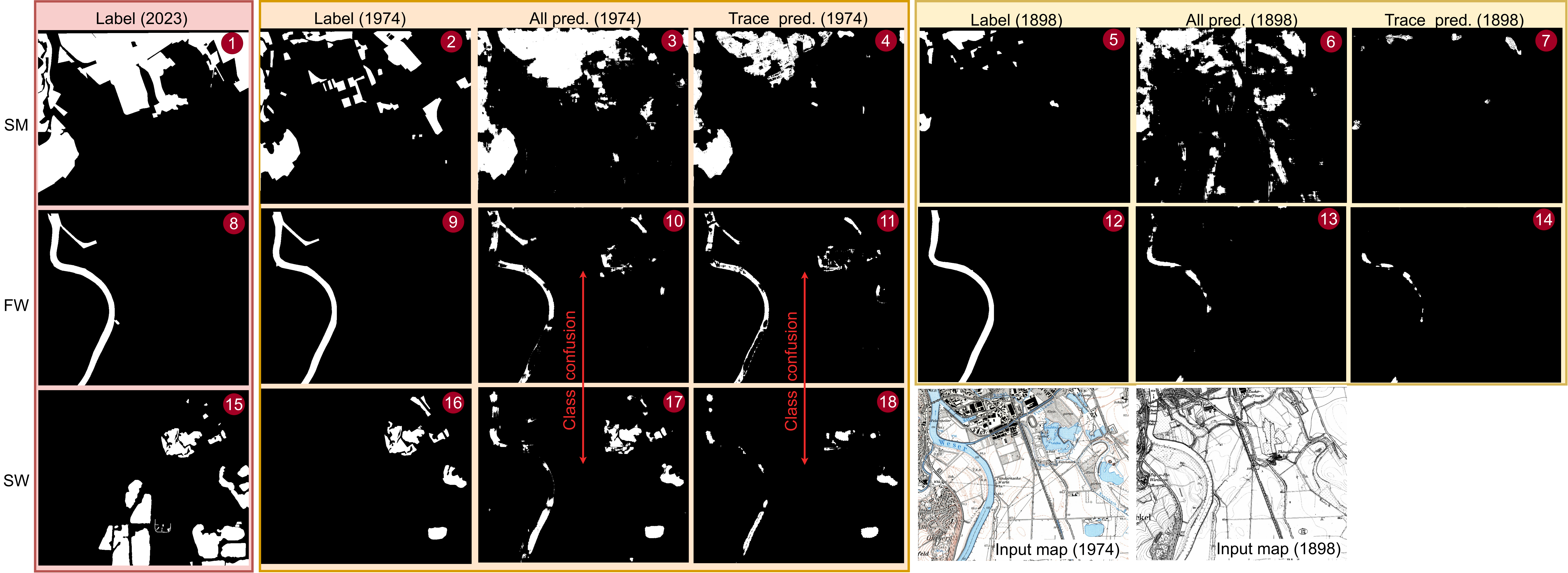}
    \caption{Exemplar result of mono-directional age-tracing for class FW, SW, SM.}
    \label{fig:examp_res}
\end{figure*}
\begin{figure*}
    \centering
    \includegraphics[width=.95\linewidth,trim={3cm 0 3cm 0},clip]{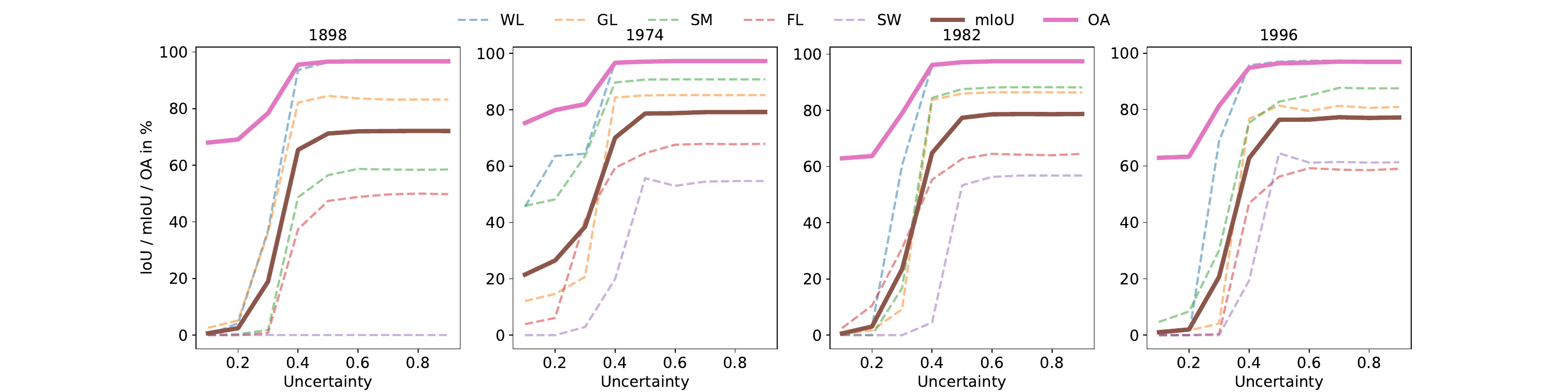}
    \caption{Semantic segmentation performance with different uncertainty thresholds for pseudo labels of bi-directional age-tracing.}
    \label{fig:bi_pseudo_unc}
\end{figure*}
\begin{figure*}
    \centering
    \includegraphics[width=.95\linewidth,trim={3cm 0 3cm 0},clip]{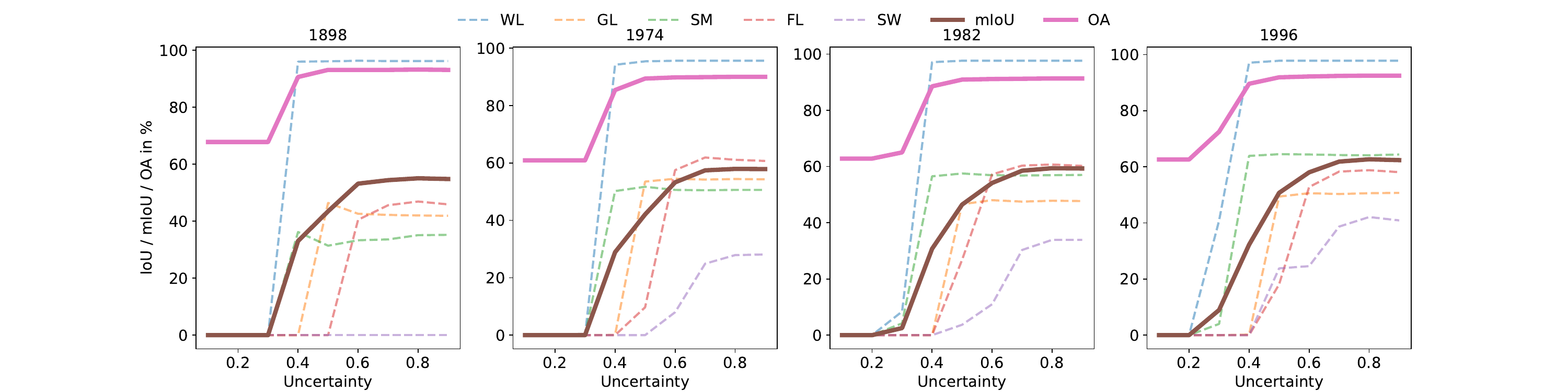}
    \caption{Semantic segmentation performance with different uncertainty thresholds for pseudo labels of mono-directional age-tracing.}
    \label{fig:mono_pseudo_unc}
\end{figure*}

\section{Result and evaluation}
The quantitative results of semantic segmentation results are shown in \cref{tab:1898} to \cref{tab:1996}. Note that the Standing Water (SW) class is not available in the map of 1898; hence, it is not accounted for in the evaluation.

\subsection{Bi-directional age-tracing}

Across all ages, the  model $Trace_{bi}$ that is trained with our proposed age-tracing method has the best overall $mIoU$ and $OA$ performance. Compared to the baseline models $Pre_{bi}$ and $All_{bi}$, it has significant improvement which is reflected, especially, by the $mIoU$ metric. From year 1898 to 1996, the $mIoU$ of $Trace_{bi}$ has increased about $1\%$, $10\%$, $16\%$,  $20\%$ on top of $Pre_{bi}$. This validated the assumption that the model can gain knowledge from the similarities between the historical maps with small age derivations.
The performance gaps between $Trace_{bi}$ and the baseline models are getting larger as the age difference to the anchor year $a_i$ increases. Because the performance of $Trace_{bi}$ is staying relatively stable while the baseline models perform much worse as the age difference increases. From year 1974 to 1966, the $mIoU$ of $Trace_{bi}$ only decreased about $3\%$ while $Pre_{bi}$ and $All_{bi}$ have decreased about $10\%$ and $16\%$, respectively. However, the result in 1989 is not consistent with this conclusion because of the unavailability of the SW class. The $IoU$ of this class is decreasing as the age difference increases. Reasonably, the $mIoU$ increases without the presence of this low-performing class for computing the average. 

\subsection{Mono-directional age-tracing}

Compared to bi-directional, the mono-directional age-tracing is more challenging because of the longer tracing path from 2023 to 1898. Moreover, the historical maps are only available till 2017 while the labels are only available in 2023. This age gap leads to an inconsistency of land use coverage, hence also deteriorates the model performance while training with such error-prone labels. As the age traces back, the inconsistent areas are also expanding. As shown in \cref{fig:incos}, the labels have great inconsistency between 1974 and 2023, especially for class GL and SW. As a result, these challenges have led to overall worse performance in comparison to the bi-directional tracing. Because of the large domain gap, the model trained on historical maps from 2016 and 2017 can hardly be adapted to the maps in other ages. For instance, the $Pre_{mono}$ is barely predicting on maps of 1898. With all maps across all ages and the label only from 2023 for training, $All_{mono}$ performs much better than $Pre_{mono}$ because $All_{mono}$ has seen all the maps to close the domain gap. However, because of lower consistency between labels and input maps, $All_{mono}$ performs worse than $All_{bi}$. By age-tracing, the $OA$s are increased, indicating that more pixels are correctly classified. More specifically, this improvement is mainly contributed by the WL and GL class whose $IoU$s are increased compared to the baseline models.

However, because of the deteriorating performance of the minority classes (\eg~FW, SW), the $mIoU$s of $Trace_{mono}$ are pulled down compared to $All_{mono}$. This is caused by the poor performance of the model on distinguishing between the two water classes $FW$ and $SW$. It can be observed in \cref{fig:examp_res}: $All_{mono}$ and $Trace_{mono}$  are confused about the class FW and SW (\textit{class confusion}) and generate mixed labels for each of these two classes. Using these error-prone pseudo labels, the $Trace_{mono}$ performs worse than $All_{mono}$ that uses the label of 2023 which has better consistency with the data, especially for the class FW (\cref{fig:examp_res} \textcircled{8} vs. \textcircled{9}). Differently, the $IoU$ of class SM has slight improvement in 1982 and 1996 by age-tracing, but worse performance in 1898 and 1974. We think this is because the SM class is shrinking to a minority class as age traces back. This shrinkage can be observed by comparing the SM label in 2023 (\cref{fig:examp_res} \textcircled{1}) and 1898 (\cref{fig:examp_res} \textcircled{5}), it can lead the model biased to classifying the target pixels to the majority class.

\subsection{Uncertainty threshold for pseudo labels}

To find the best uncertainty threshold for generating the pseudo labels, we configure this threshold with different values from $0.1$ to $0.9$ and train one  $Trace_{bi}$ and one $Trace_{mono}$ with each of these values. The results are shown in \cref{fig:bi_pseudo_unc} and \cref{fig:mono_pseudo_unc}. Dashed lines are the $IoU$s for each class and the bold solid lines are the overall $mIoU$s and $OA$s. All lines have an overall increasing tendency as the uncertainty threshold increases. Higher uncertainty indicates lower confidence for the classification results, hence more pixels under the high uncertainty threshold are selected as pseudo labels for training. For both bi- and mono-directional age-tracing, the $OA$s have significant improvements as the uncertainty threshold increases from $0.1$ to $0.5$ and then reaches a plateau. 

In comparison, the $mIoU$s reach a plateau at a higher threshold, especially for the mono-directional age-tracing which is at about $0.8$ (brown bold line). This is because the $IoU$s are more sensitive to the biased data  of unbalanced sample ratio for different classes, especially when inaccurate pseudo labels are used for training. For instance in \cref{fig:bi_pseudo_unc}, the $IoU$ of the minority class FW and SW in 1982 reaches a plateau at the uncertainty threshold of about 0.6 while that of majority classes WL and GL happens at about 0.4. With more inaccurate pseudo labels for mono-directional age-tracing, this plateau threshold gap in 1892 gets larger, about 0.8 for the two minority water-related classes and still 0.4 for the two majority classes.

Although the overall performance finally reaches a plateau that indicates a stable performance at high uncertainty thresholds, there are still small fluctuations as shown in \cref{tab:pseudo_unc}. The bold texts in the table indicate the best performed models. On average, an uncertainty threshold of about $0.8$ has the best performance. This value is recommended to train the model with the age-tracing strategy.

\section{Conclusion}
In this paper, we propose an age-tracing strategy that leverages the temporal similarities between historical maps from neighboring years to train a generic semantic segmentation model capable of segmenting maps across all ages. The strategy begins by pre-training a model on a labeled anchor year, then traces map ages both backward and forward for fine-tuning until reaching the earliest and latest maps in the dataset. Experiments were conducted on the newly curated \textit{Hameln} dataset, using two different anchor years: one from the middle (1973-1975) and one from the end (2023) of the entire age range (1897-2017). The results demonstrate that age-tracing improves overall segmentation accuracy compared to baselines trained solely on maps from the anchor year or across all available ages.
Additionally, we experimented with various parameter configurations for the uncertainty thresholds used to generate pseudo-labels, confirming that higher uncertainty (\eg~0.8) yields the best performance improvement.

For future work, additional configurations—such as input image size and augmentation techniques—can be explored. Furthermore, we observed that the model struggles to distinguish between the "flowing water" and "standing water" classes. This issue may be due to the limited receptive field of the model, which could be addressed in future research.

\begin{table}[t!]
    \centering
\resizebox{.5\textwidth}{!}{ %
    \begin{tabular}{c|c|c|c|c|c}\toprule
     unc. & 0.5 & 0.6 & 0.7 & 0.8 & 0.9\\\hline
        \rowcolor{lightgray} \multicolumn{6}{c}{mIoU/OA (bi-directional)} \\\hline
    1898 & 71.2/96.6 & 72.0/96.7 & \textbf{72.1/96.7} & \textbf{72.1/96.7} & \textbf{72.1/96.7}\\\hline
    1974 & 78.7/97.1 & 78.8/97.3 & \textbf{79.2/97.3} & \textbf{79.2/97.3} & \textbf{79.2/97.3}\\\hline
    1982 & 77.3/97.1 & \textbf{78.6/97.4} & 78.6/97.4 & 78.6/97.4 & 78.6/97.4\\\hline
    1996 & 76.4/96.4 & 76.4/96.6 & \textbf{77.3/97.0} & 77.0/96.9 & 77.2/96.9\\\hline
      \rowcolor{lightgray} \multicolumn{6}{c}{mIoU/OA (mono-directional)} \\\hline
     1898 & 43.5/93.1 & 53.2/93.1 & 54.4/93.1 & \textbf{55.0/93.2} & 54.8/93.1\\\hline
     1974 & 42.1/89.4 & 53.2/89.8 & 57.4/89.9 & \textbf{57.9/90.0} & \textbf{57.9/90.0}\\\hline
     1982 & 46.4/90.9 & 54.1/91.1 & 58.5/91.2 & \textbf{59.4/91.3} & 59.3/91.3\\\hline
     1996 & 50.7/91.9 & 58.0/92.2 & 61.9/92.4 & \textbf{62.7/92.5} & 62.4/92.5\\
     \bottomrule
    \end{tabular}
} %
    \caption{Model performance with different uncertainty thresholds. Bold text indicates the best performed models.}
    \label{tab:pseudo_unc}
\end{table}


\begin{spacing}{1.0}
      \bibliography{ICAguidelines_bib} 

\begin{thebibliography}{xx}

\bibitem[Bromberg and Bertness, 2005]{salt_marsg}
Bromberg, K.~D. and Bertness, M.~D., 2005.
\newblock Reconstructing new england salt marsh losses using historical maps.
\newblock {\em Estuaries} 28, pp.~823--832.

\bibitem[Csurka et al., 2023]{Csurka2023SemanticIS}
Csurka, G., Volpi, R. and Chidlovskii, B., 2023.
\newblock Semantic image segmentation: Two decades of research.
\newblock {\em Found. Trends Comput. Graph. Vis.} 14, pp.~1--162.

\bibitem[Ekim et al., 2021]{ekim_automatic_2021}
Ekim, B., Sertel, E. and Kabaday?, M.~E., 2021.
\newblock Automatic {Road} {Extraction} from {Historical} {Maps} {Using} {Deep} {Learning} {Techniques}: {A} {Regional} {Case} {Study} of {Turkey} in a {German} {World} {War} {II} {Map}.
\newblock {\em ISPRS International Journal of Geo-Information} 10(8), pp.~492.

\bibitem[Farahani et al., 2021]{da}
Farahani, A., Voghoei, S., Rasheed, K. and Arabnia, H.~R., 2021.
\newblock A brief review of domain adaptation.
\newblock In: R.~Stahlbock, G.~M. Weiss, M.~Abou-Nasr, C.-Y. Yang, H.~R. Arabnia and L.~Deligiannidis (eds), \emph{Advances in Data Science and Information Engineering}, Springer International Publishing, Cham, pp.~877--894.

\bibitem[Heitzler and Hurni, 2020]{heitzler_cartographic_2020}
Heitzler, M. and Hurni, L., 2020.
\newblock Cartographic reconstruction of building footprints from historical maps: {A} study on the {Swiss} {Siegfried} map.
\newblock {\em Transactions in GIS} 24(2), pp.~442--461.
\newblock \_eprint: https://onlinelibrary.wiley.com/doi/pdf/10.1111/tgis.12610.

\bibitem[Levin et al., 2010]{Levin_Kark_Galilee_2010}
Levin, N., Kark, R. and Galilee, E., 2010.
\newblock Maps and the settlement of southern palestine, 1799-1948: an historical/gis analysis.
\newblock {\em Journal of Historical Geography} 36(1), pp.~1--18.

\bibitem[Leyk, 2010]{leyk_segmentation_2010}
Leyk, S., 2010.
\newblock Segmentation of {Colour} {Layers} in {Historical} {Maps} {Based} on {Hierarchical} {Colour} {Sampling}.
\newblock In: J.-M. Ogier, W.~Liu and J.~Llad?s (eds), \emph{Graphics {Recognition}. {Achievements}, {Challenges}, and {Evolution}}, Springer, Berlin, Heidelberg, pp.~231--241.

\bibitem[Ojala et al., 2002]{lbp}
Ojala, T., Pietikainen, M. and Maenpaa, T., 2002.
\newblock Multiresolution gray-scale and rotation invariant texture classification with local binary patterns.
\newblock {\em IEEE Transactions on Pattern Analysis and Machine Intelligence} 24(7), pp.~971--987.

\bibitem[Picuno et al., 2019]{picuno_investigating_2019}
Picuno, P., Cillis, G. and Statuto, D., 2019.
\newblock Investigating the time evolution of a rural landscape: {How} historical maps may provide environmental information when processed using a {GIS}.
\newblock {\em Ecological Engineering} 139, pp.~105580.

\bibitem[Ronneberger et al., 2015]{unet}
Ronneberger, O., Fischer, P. and Brox, T., 2015.
\newblock U-net: Convolutional networks for biomedical image segmentation.
\newblock In: N.~Navab, J.~Hornegger, W.~M. Wells and A.~F. Frangi (eds), \emph{Medical Image Computing and Computer-Assisted Intervention -- MICCAI 2015}, Springer International Publishing, Cham, pp.~234--241.

\bibitem[San Antonio~Gómez et al., 2014]{san-antonio-gomez_urban_2014}
San Antonio~Gómez, C., Velilla, C. and Manzano~Agugliaro, F., 2014.
\newblock Urban and landscape changes through historical maps: {The} {Real} {Sitio} of {Aranjuez} (1775-2005), a case study.
\newblock {\em Computers, Environment and Urban Systems} 44, pp.~47--58.

\bibitem[Shelhamer et al., 2014]{Shelhamer2014FullyCN}
Shelhamer, E., Long, J. and Darrell, T., 2014.
\newblock Fully convolutional networks for semantic segmentation.
\newblock {\em 2015 IEEE Conference on Computer Vision and Pattern Recognition (CVPR)} pp.~3431--3440.

\bibitem[Tonolla et al., 2021]{tonolla_seven_2021}
Tonolla, D., Geilhausen, M. and Doering, M., 2021.
\newblock Seven decades of hydrogeomorphological changes in a near-natural ({Sense} {River}) and a hydropower-regulated ({Sarine} {River}) pre-{Alpine} river floodplain in {Western} {Switzerland}.
\newblock {\em Earth Surface Processes and Landforms} 46(1), pp.~252--266.
\newblock \_eprint: https://onlinelibrary.wiley.com/doi/pdf/10.1002/esp.5017.

\bibitem[Uhl et al., 2020]{uhl_automated_2020}
Uhl, J.~H., Leyk, S., Chiang, Y.-Y., Duan, W. and Knoblock, C.~A., 2020.
\newblock Automated {Extraction} of {Human} {Settlement} {Patterns} {From} {Historical} {Topographic} {Map} {Series} {Using} {Weakly} {Supervised} {Convolutional} {Neural} {Networks}.
\newblock {\em IEEE Access} 8, pp.~6978--6996.
\newblock Conference Name: IEEE Access.

\bibitem[Uhl et al., 2021]{Uhl_Leyk_2021}
Uhl, J.~H., Leyk, S., Li, Z., Duan, W., Shbita, B., Chiang, Y.-Y. and Knoblock, C.~A., 2021.
\newblock Combining remote-sensing-derived data and historical maps for long-term back-casting of urban extents.
\newblock {\em Remote Sensing} 13(1818), pp.~3672.

\bibitem[Wu et al., 2022a]{wu_closer_2022}
Wu, S., Heitzler, M. and Hurni, L., 2022a.
\newblock A {CLOSER} {LOOK} {AT} {SEGMENTATION} {UNCERTAINTY} {OF} {SCANNED} {HISTORICAL} {MAPS}.
\newblock {\em The International Archives of the Photogrammetry, Remote Sensing and Spatial Information Sciences} XLIII-B4-2022, pp.~189--194.

\bibitem[Wu et al., 2022b]{wu_leveraging_2022}
Wu, S., Heitzler, M. and Hurni, L., 2022b.
\newblock Leveraging uncertainty estimation and spatial pyramid pooling for extracting hydrological features from scanned historical topographic maps.
\newblock {\em GIScience \& Remote Sensing} 59(1), pp.~200--214.

\bibitem[Wu et al., 2023]{wu_domain_2023}
Wu, S., Schindler, K., Heitzler, M. and Hurni, L., 2023.
\newblock Domain adaptation in segmenting historical maps: {A} weakly supervised approach through spatial co-occurrence.
\newblock {\em ISPRS Journal of Photogrammetry and Remote Sensing} 197, pp.~199--211.

\bibitem[Yuan et al., 2023]{Yuan_gevbev2023}
Yuan, Y., Cheng, H., Yang, M.~Y. and Sester, M., 2023.
\newblock Generating evidential bev maps in continuous driving space.
\newblock {\em ISPRS Journal of Photogrammetry and Remote Sensing} 204, pp.~27--41.

\end{thebibliography}
\end{spacing}

\label{LastPage} 
\end{document}